\documentclass[journal]{IEEEtran}
\usepackage{lineno,hyperref}
\modulolinenumbers[5]
\usepackage{amsmath}
\usepackage{amssymb}
\usepackage{booktabs}
\usepackage{graphicx}
\usepackage{bm}

\ifCLASSINFOpdf
\else
\fi
\begin{document}

%
\title{Attention-Based 3D Seismic Fault Segmentation Training by a Few 2D Slice Labels}
%
%
%

\author{Yimin Dou,
        Kewen Li,
        Jianbing Zhu,
        Xiao Li,
        Yingjie Xi,
        
\thanks{The corresponding author is Kewen Li. likw@upc.edu.cn}
\thanks{Yimin Dou ,Kewen Li, Xiao Li and Yingjie Xi, College of computer science and technology, China University of Petroleum (East China) Qingdao, China.}
\thanks{Jianbing Zhu, Geophysical Research Institute Shengli Oilfield Company, SINOPEC Dongying, China.}
\thanks{This work was supported by grants from the National Natural Science Foundation of China (Major Program, No.51991365).}
}

\maketitle

\begin{abstract}
Detection faults in seismic data is a crucial step for seismic structural interpretation, reservoir characterization and well placement. Some recent works regard it as an image segmentation task.
The task of image segmentation requires huge labels, especially 3D seismic data, which has a complex structure and lots of noise. Therefore, its annotation requires expert experience and a huge workload.
In this study, we presented $\lambda$-BCE and $\lambda$-smooth $L_1$ loss to effectively train 3D-CNN by some slices from 3D seismic volume label, so that the model can learn the segmentation of 3D seismic data from a few 2D slices. In order to fully extract information from limited data and suppress seismic noise, we proposed an attention module that can be used for active supervision training and embedded in the network. The attention map label is generated by the original label, and letting it supervise the attention module using the $\lambda$-smooth $L_1$ loss. The experimental results demonstrate the proposed loss function can extract 3D seismic features from a few 2D slice labels. 
And it also shows the advanced performance of the attention module, which can significantly suppress the noise in the seismic data while increasing the sensitivity of the model to the foreground.
Finally, on the public test set, the proposed method achieved similar performance to using 3D volume labels by using only 3.3\% of the slices.

\end{abstract}

\begin{IEEEkeywords}
Seismic fault detection, Interpretation, Seismic attributes 
\end{IEEEkeywords}

%
\IEEEpeerreviewmaketitle

\section{Introduction}

The formation mechanism of fault is complex and its types are diverse, furthermore fault detection is a crucial step for seismic structural interpretation, reservoir characterization and well placement, which determines that fault detection is an important topic in the field of oil-gas exploration. The main methods of fault detection include seismic methods and imaging logging method  \cite{29,30,31}, and our study focuses on the use of seismic data for fault detection.

Before deep learning was widely used, researchers used traditional geological methods for fault detection. The first to be applied to fault detection was the theory of   fault anisotropy. Crampin discovered and put forward many new understandings and opinions on fault anisotropy \cite{1}, Rüger proposed the Rüger approximation formula, which verified that the formula has good adaptability in weakly anisotropic media \cite{2}, and proposed AVO (Amplitude Variation with Offset) gradient inversion to calculate the fault parameters \cite{3}, but the anisotropy  as a basic property of the fault, is prone to noise interference from seismic data when applied to the detection task, and the detection accuracy is very low; Bahorich proposed the use of correlation volume detection of seismic faults \cite{4}, by calculating the cross-correlation coefficients between seismic traces to highlight the characteristics of fault discontinuity, but for seismic data with relatively large coherent noise or small faults, the detection effect is poor. Subsequently, Marfurt et al. proposed the second-generation coherent volume technology, which improved the anti-noise ability, but the resolution was low \cite{5}. The third-generation coherent technology provided high resolution detection results in noisy data by calculating the eigenvalues of covariance matrix, but the effect was not good in some special geological environment \cite{6} (such as the wing of salt mound).

Pedersen applied ant colony algorithm to fault detection \cite{7}, ant colony algorithm used ant tracking to highlight fault lines, filtered irrelevant noise and non-fault response; Sun et al. combined spectral decomposition technology with ant colony algorithm \cite{8}; Aqrawi uses the improved 3D Sobel filtering method and ant colony algorithm to realize the detection of small faults \cite{9}. However, in the 3D seismic data fault detection, the ant colony algorithm interpretation of the fault in inline and crossline is very rough, and the detection results from the timeline do not conform to the characteristics of the fault distribution. In addition, there are some other fault detection algorithms. Saito and Hayashi use frequency domain-based Stoneley waves to detect faults \cite{10}; Admasu et al. proposed an active contour algorithm to achieve semi-automatic tracing of faults \cite{11}; Priezzhev and scollard proposed a fault detection method based on orthogonal decomposition of seismic data \cite{12}; Hale uses three steps: calculation of 3D tomographic images, extraction of fault surface and drop estimation to detect faults \cite{13}; Wang et al. proposed based on Hough transform and Vector tracking interactive fault detection algorithm \cite{14}; Wu and Fomel proposed a method to extract the optimal surface based on the maximum fault attributes, and use these optimal surface voting to detect faults \cite{15}. However, the use of traditional geological methods or the introduction of digital image processing algorithms on this basis cannot solve the problem of strong noise interference in seismic data.

As early as 2005, Tingdahl et al. realized an algorithm that uses multiple seismic attributes and multilayer perceptron (MLP) to detect faults \cite{16}, but its performance is limited by the neural network theory and hardware conditions at that time. With the development of deep learning in recent years, some studies have introduced convolutional neural networks into seismic fault detection \cite{17,18,19,20}. These methods regard fault detection as an image segmentation task in the field of computer vision.   2D seismic image pixels are classified into two categories (fault and non-fault), but doing so will lose the 3D spatial morphological feature of the fault, which will cause discontinuities in the segmentation results of the faults of adjacent slices, in other words, doing segmentation on inline will cause discontinuities on crossline (Figure \ref{fig10}, \ref{fig11}). Guitton uses 3D CNN for fault segmentation \cite{21}, instead of choosing the currently widely used pixel level segmentation structure \cite{ronneberger2015u,shelhamer2017fully}, his method is to divide the seismic data into cuboids and classifies each one, which cannot learn the spatial structure of seismic data. The label uses the result of the algorithm proposed by Hale \cite{13} as the label, and the upper limit of its model performance depends on the result of Hale's algorithm; Wu et al. use synthetic seismic data is used to train the 3D U-Net model \cite{22,23}. Synthetic data avoids problems caused by manual labeling. However, in many cases, synthetic data cannot be generalized to real seismic data. We verified the work of Wu et al. There is still a lot of noise in the prediction results of the model trained on synthetic data on the real data we provide (Figure \ref{fig5}), which is difficult to apply in the field of petroleum exploration with various types of geological structures.

In short, the segmentation of 3D faults still faces two major problems. First of all, various complex geological conditions and the influence of acquisition equipment lead to a low signal-to-noise ratio of the original seismic data, resulting in a large amount of noise in the detection results obtained by traditional geological methods or machine learning methods; secondly, 3D seismic data cannot be directly labeled, the workload of labeling on 2D data and then synthesizing 3D labels is huge and requires expert experience. And the adjacent slices of the seismic fault are very similar, redundant labels that consume a lot of labor cost may not significantly improve the training performance.

In this study, we propose an attention module and embed it in U-Net to suppress noise in seismic data, and propose the $\lambda$-BCE and $\lambda$-smooth $L_1$ loss so that the 3D segmentation model can be trained by a few 2D slice labels.

At present, most of the work on the attention mechanism makes the model adaptively generate the attention area \cite{24,32,33,34}, that is, the attention map, which suppresses the background area through the attention map.
However, such methods are usually insensitive to small targets \cite{24}, especially when the foreground and background of seismic fault data are seriously out of balance, and the label pixels of the fault are usually arranged in a line, and the adaptive attention mechanism is difficult to capture the key features and area of the fault. Different from other methods, the attention map of our proposed attention module is generated by active supervised learning, so it is called active attention module (AAM). Although a supervised learning method is adopted, we use Gaussian function to simulate the weight distribution of attention, and then combine with the original ground truth to generate the supervision information of the module, so no additional manual labeling is required. And this module adds additional supervision information to the training, which can be regarded as an intermediate supervision mechanism \cite{25}, which can improve the performance of the model even in synthetic data with few noise. Experiments show that this module can significantly improve model segmentation performance and suppress noise.

Çiçek uses a combination of softmax and cross-entropy loss function to train a medical image segmentation model using sparsely labeled data \cite{23}, and treats unlabeled voxels as additional categories, which brings redundant neurons and model parameters. We extend the approach of sparse labeling to the BCE (Binary Cross-Entropy, equation (\ref{bce})) loss. The difference between BCE and cross-entropy is that BCE is used for binary classification problems, which express probabilities through a single value, while cross-entropy is used for multi-category problems, which usually express probabilities using multiple values, and the category corresponding to the largest value is taken as the final category. We improve the BCE by proposing $\lambda$-BCE loss, which can complete the training of 3D segmentation models under sparse labels without setting additional categories, avoiding redundant parameters and computational effort. Moreover, the $\lambda$-smooth $L_1$ loss is proposed using the same way to improve the smooth $L_1$ \cite{26}. We use them and AAM for seismic fault segmentation, and verified the effectiveness of this method and tested the effects of multiple labeling modes on model performance through experiments. We found that under our method, using a few 2D slices training can achieve model performance similar to using 3D volume labels.

In summary, we propose an attention module that can be actively supervised and trained (Active Attention Module, AAM) based on the characteristics of seismic faults, its supervision information is generated by the combined simulation of ground truth and Gaussian function, and which generated attention map is used to suppress the seismic noise. In addition, this module can be treated as intermediate supervision to provide more effective gradients for training, thereby increasing the sensitivity to the foreground in the back propagation to improve the performance of the model. Through the $\lambda$-BCE and $\lambda$-smooth $L_1$ loss proposed in this paper, only a few 2D slices are needed to train the 3D segmentation model, and the performance of the model under different slice labeling modes is verified through experiments. This allows geologists and oil-gas prospectors to label a few 2D slices in the seismic data to obtain accurate 3D fault segmentation models for all similar geological types of seismic data.

\section{Approach}
At present, there are two main problems with the fault segmentation of seismic data. One is that interference factors such as a large amount of noise in seismic data will affect the performance of segmentation. The second is that it is very difficult to produce fault labels for 3D seismic data, requiring an enormous  workload and expert experience.
This section discusses these two problems separately and proposes solutions. First, AAM is proposed based on the characteristics of seismic fault labels, so that the model focuses on the fault area and suppresses noise. Secondly,  $\lambda$-BCE and $\lambda$-smooth $L_1$ loss function is proposed, so that the 3D segmentation model can be trained with a few 2D slice labels.
\subsection{Active Attention Module}
The widely used segmentation models all merge with  high-level and low-level features \cite{ronneberger2015u,shelhamer2017fully}. The main reason is that low-level features contain richer edges and detailed features, while high-level features have more semantic features \cite{shelhamer2017fully}, and the fusion of the two is conducive to finer pixel classification. However, there is a strong noise in the low-level features in seismic data, which results in the introduction of not only edge features but also noise in feature fusion. In this section, we propose an attention module that can be trained through active supervision (Active Attention Module, AAM). This attention module suppresses low-level noise and makes the model pay more attention to the fault area. Unlike the way most of the current attention modules adaptively generate attention maps \cite{24,32,33,34}, our attention maps are generated through  supervised training.

This module obtains the linear projections $\omega_l$$F_l$ and $\omega_h$$F_h$ from the low-level detail feature $F_l$ and the high-level semantic feature $F_h$ through $1\times1$ convolution respectively, and then combines them into a single channel. The whole process is expressed as,
\begin{equation}
	\hat{\Theta}=\omega_s\mbox{ReLU}^T(\omega_lF_l+\omega_hF_h)\label{attetion}
\end{equation}
where $\omega_l$ and $\omega_h$ is differentiable, so in equation (\ref{attetion}), $\omega_lF_l+\omega_hF_h$ (denoted as $F_s$) can be interpreted as the difference between the low-level features and the high-level features. With the deepening of network layers, the features extracted from the deep layer will trend to the ground truth increasingly \cite{25}. Therefore, $F_s$ shows the signal response of ground truth. The $\hat{\Theta}$ that is weighted by $F_s$ is the attention map we need. 

Many work related to attention mechanism hopes to adaptively capture the attention map, but it is difficult to achieve on small objects. Especially in seismic fault segmentation task, the foreground and background are seriously out of balance, and the adaptive attention mechanism is difficult to work. Therefore, we propose a method to generate attention map label (denoted as $\Theta$) from ground truth to supervise $\hat{\Theta}$ to generate attention map.

We hope that the attention map can effectively suppress the non-fault area feature  in $F_l$, and retain the feature of the fault area. So on the idealized attention map, the closer to the fault area, the greater the weight, and the farther from the fault area, the smaller the weight. With the decrease of euclidean distance from the fault area, the weight should be from 0 to 1. We use the Gaussian function to simulate this trend. 
Let $\bm{\mbox{x}} \in \mathbb{R}^2$ be the ground truth position of fault voxel in the data.
The value at location $\theta \in \mathbb{R}^2$ in $\Theta$ defined as,
\begin{equation}
	\Theta(\theta)=\mbox{exp}(-\frac{\Arrowvert\theta- \bm{\text{x}}\Arrowvert^2_2}{\sigma^2})\label{att_fumula}
\end{equation}
where $\sigma$ controls the spread of the peak.

\begin{figure}[ht]
	\centering

	\includegraphics[scale=0.35]{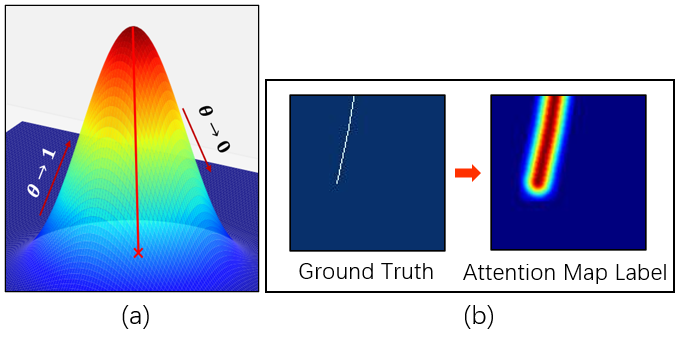}
	\centering\caption{(a) shows the trend of $\theta$. (b) shows the attention map label generated when we replace each pixel in ground truth with the response point in (a).}
	\label{fig1}
\end{figure}

The attention map label generated from ground truth is shown in Figure \ref{fig1}.

\begin{figure}[ht]
	\centering
	\includegraphics[scale=0.63]{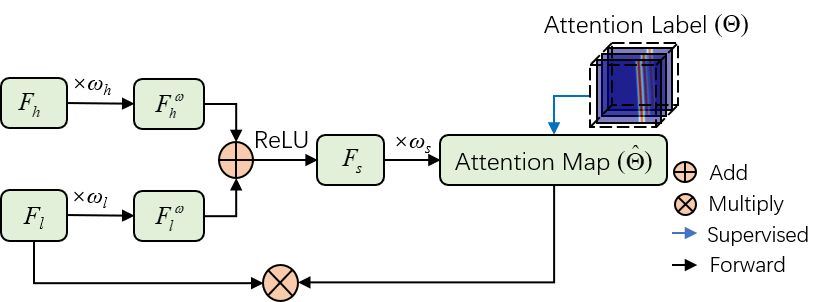}
	\centering\caption{The working principle of AAM when forward propagating is that the attention map will generate a weight approaching 1 around the fault, while the non-fault area will generate a weight approaching 0. In the decoding part of U-Net, low-level features will be fused, but it contains a strong noise. Before fusion, multiplying the attention map with low-level features can improve the sensitivity to fault areas while suppressing factors such as noise in non-fault areas.}
	\label{fig9}
\end{figure}

\begin{figure*}[ht]
	\includegraphics[scale=0.28]{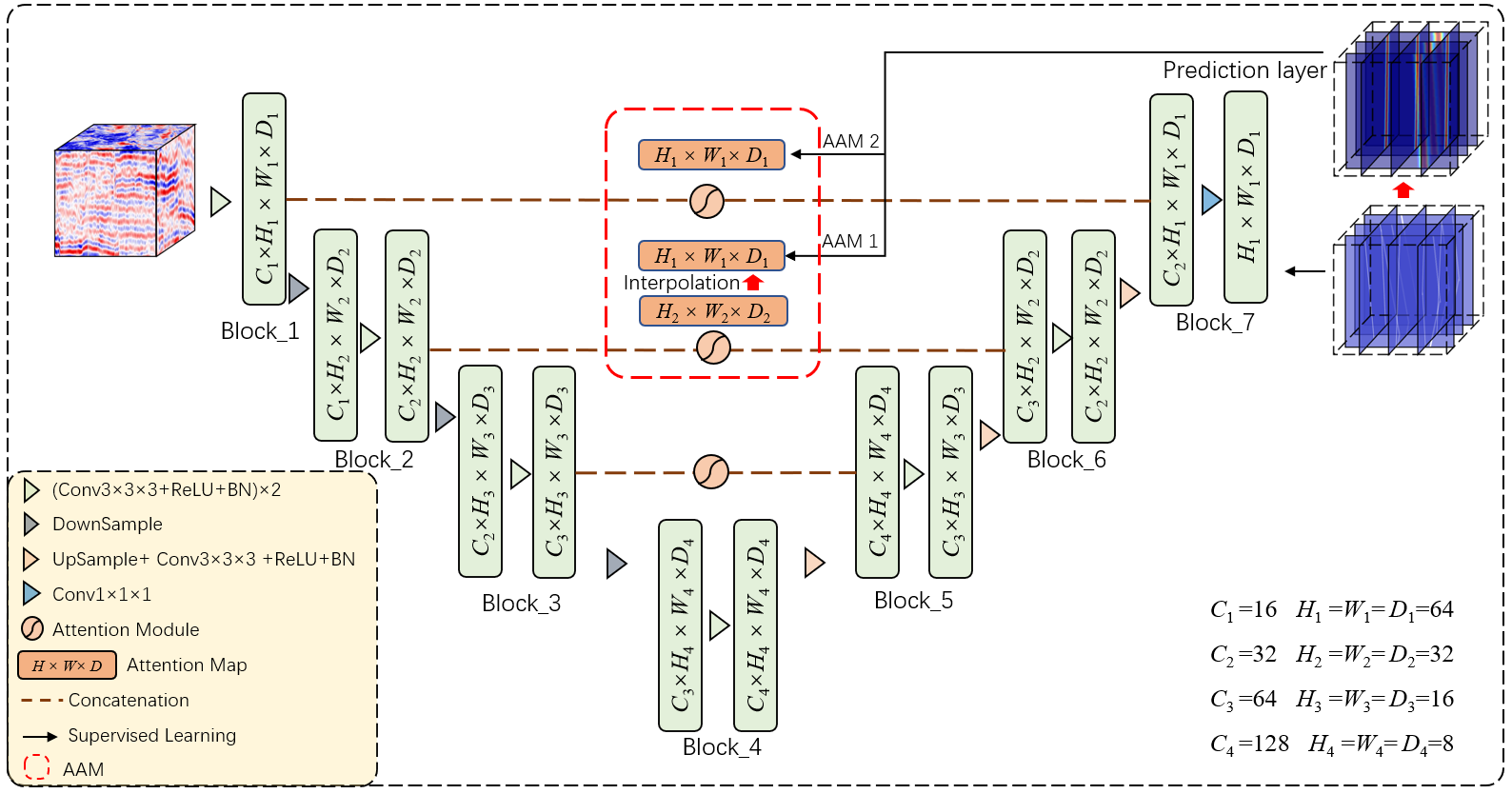}
	\centering\caption{The AAM is embedded in the basic U-Net framework, which can suppress the noise in the background region of the low-level features in the forward-propagation feature fusion phase and improve the performance by optimizing the sensitivity of the high-level and low-level features to the foreground region with additional gradients in the back-propagation. Moreover, the model is trained by our proposed $\lambda$ loss function, which is labeled with a few 2D slices, it can fully extract 3D spatial information from the limited number of slice labels, and we will describe this process in detail in the next section.}
	\label{fig2}
\end{figure*}

Use the generated attention map label to supervise $\hat{\Theta}$ through smooth $L_1$ loss \cite{26},
\begin{equation}
	\mathcal{L}_{sL_1}(\hat{\theta}_i,\theta_i)=\sum_{i=1}^{whd}\mbox{smooth}_{L_1}(\hat{\theta}_i-\theta_i)
\end{equation}
where.
\begin{equation}
	\mbox{smooth}_{L_1}(x)=\begin{cases}
		0.5x^2 \     &\mbox{if}|x|<1\\
		|x|-0.5 \    &\mbox{otherwise}
	\end{cases}
\end{equation}
The reason for using smooth $L_1$ loss is to ensure that when the model cannot extract enough features at the initial stage of training, the large difference between the predicted value and the ground truth leads to a high gradient that causes training instability. And the difference between the later predicted value and ground truth is very small and still can provide a stable gradient. The structure and principle of AAM are shown in Figure \ref{fig9}, and the overall structure of the model is shown in Figure \ref{fig2}.

AAM improves model performance in two ways. First, when propagating forward, by generating $\hat{\Theta}$ and then multiplying with $F_l$, the low-level noise features are suppressed, which is equivalent to selecting the approximate area to be segmented in advance, helping the final classification layer to filter the area in advance. Figure \ref{fig9} illustrates the working principle of forward propagation in detail.

Second, AAM is an intermediate supervision mechanism that provides an additional effective gradient in the back propagation process.
The $\Theta$ generated from the original label can be regarded as an ideal feature weight distribution map.
So $\Theta$ is not only used to supervise the generation of attention map by the AAM module, but also effectively regulates the high-level and low-level features during the training process and increases the sensitivity to foreground voxels. Wei et al. discussed the mechanism of intermediate supervision in reference \cite{25}.
Equations (\ref{back1}) and (\ref{back2}) explain the influence of AAM's intermediate supervision mechanism on $F_h$ and $F_l$ in the back propagation of the model.
\begin{equation}
	\frac{\partial\mathcal{L}_{sL_1}(\hat{\Theta},\Theta)}{\partial F_h} =	\frac{\partial\mathcal{L}_{sL_1}(\hat{\Theta},\Theta)}{\partial \hat{\Theta}}
	\frac{\partial\hat{\Theta}}{\partial F_s}
	\frac{\partial F_s}{\partial F_h^\omega}
	\frac{\partial F_h^\omega}{\partial F_h}
	\label{back1}
\end{equation}
\begin{equation}
	\frac{\partial\mathcal{L}_{sL_1}(\hat{\Theta},\Theta)}{\partial F_l} =	\frac{\partial\mathcal{L}_{sL_1}(\hat{\Theta},\Theta)}{\partial \hat{\Theta}}
	\frac{\partial\hat{\Theta}}{\partial F_s}
	\frac{\partial F_s}{\partial F_l^\omega}
	\frac{\partial F_l^\omega}{\partial F_l}
	\label{back2}
\end{equation}

In addition, most of the ground truth of seismic fault are arranged in one-dimensional lines, and the idealized attention weight distribution can be directly simulated by Gaussian function (Figure \ref{fig1}).
In natural or medical images, ground truth exists in the form of two-dimensional planes, from which it is difficult to estimate reasonable attention weight distribution. Therefore, if their weight distribution can be estimated by a reasonable method, this attention module may be transferred to other segmentation tasks.

AAM filters the background feature during the feature fusion process of the model and retains the key areas of low-level features.
Moreover, as an intermediate supervision mechanism, it can provide additional gradients during training, and in back propagation, the performance is improved by optimizing the sensitivity of high-level and low-level features to the foreground area.

\subsection{Learning 3D segmentation from few 2D labeled seismic data slices}
labeling the fault of 3D seismic data is a huge workload and requires expert experience. Because adjacent seismic slices are often similar, traditional data labeling methods will bring a lot of redundant labels, and manual labeling of 3D faults on 2D slices is difficult to ensure the continuity of fault labels in three-dimensional space.

This section proposes $\lambda$-BCE and $\lambda$-smooth $L_1$ loss and the sparse fault labeling method, which requires only a few slices to train 3D seismic data, avoids redundant labeling, and greatly reduces the workload of data labeling. Because the labeling is sparse, the confidence of the unlabeled area is automatically generated by the model, which solves the problem of discontinuity of artificial labels in three-dimensional space.

\begin{figure}[ht]
	\centering \includegraphics[scale=0.2]{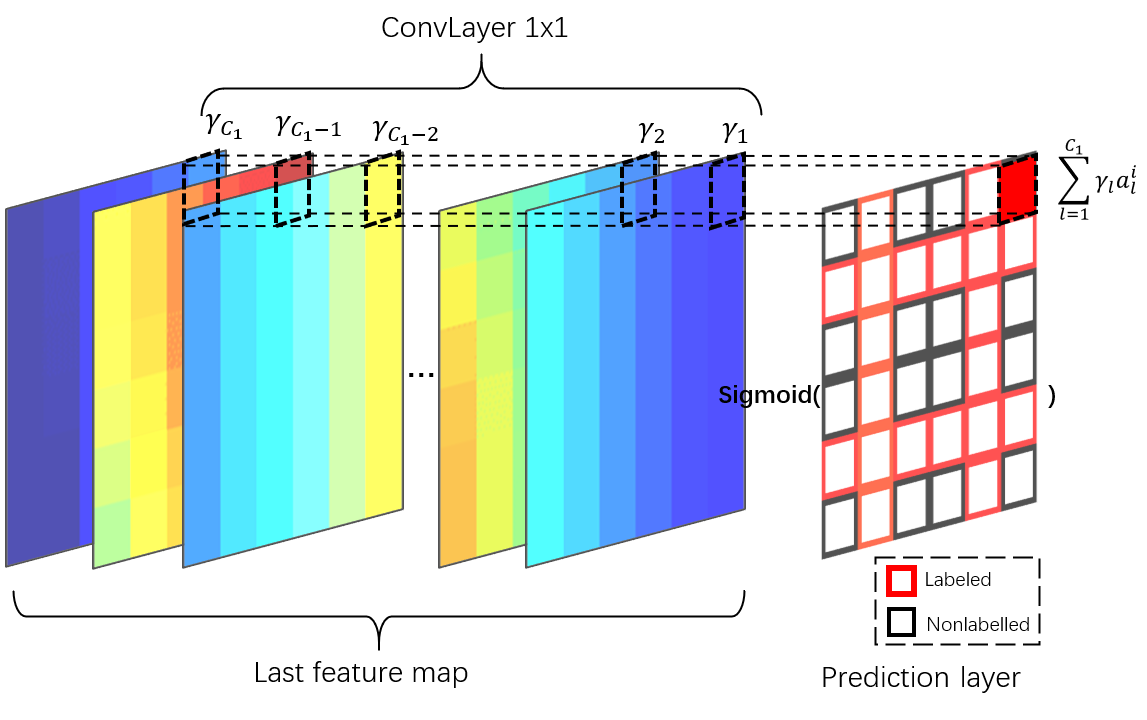}
	\caption{The elements on the last feature map share a set of weights, which allows us to obtain effective gradients only by training using the labeled voxels in the label.}
	\label{fig3}
\end{figure}

U-Net can be divided into two parts, backbone and prediction layer. The backbone part is used to extract features. The prediction layer is one convolution layer. Use $\gamma_l, l\in[1,C_1]$ to represent the convolution kernel of this layer, then the shape of $\gamma_l$ is $(C_1,k,k,k,C_0)$, where $C_1$ is the number of channels in the upper layer, $k=1$, is the size of the convolution kernel, $C_0=1$, is the number of convolution kernels, $C_1\times k\times k\times k\times C_0=C_1$, so $\gamma_l$ forms a vector of length $C_1$. As shown in Figure \ref{fig3}, for the convenience of presentation, the figure shows a two-dimensional situation, where the last feature map shares a set of convolution weights $\gamma_l$, and the weight slides on the last feature map to obtain the final result of the prediction $\mathcal{P}$, which can be expressed as equation (\ref{f6}).
\begin{equation}
	\mathcal{P} = \mbox{sigmoid}(\{\sum_{l=1}^{C_1}\gamma_la^1_l,\sum_{l=1}^{C_1}\gamma_la^2_l,...\sum_{l=1}^{C_1}\gamma_la^{whd}_l \})\label{f6}
\end{equation}
where,
\begin{equation}
	\mbox{sigmoid}(x)=\frac{1}{e^{-x}+1} 
\end{equation}
the $a_l^i$ represents the value of each element on the last feature. The label in Figure \ref{fig3} is sparse, that is, only the red part is labeled. Our method is to calculate only the gradient caused by the labeled area in backward. The last feature map shares the convolution weight, so even if some voxels that are not labeled are missing, it can still provide an effective gradient in backward. The main process is as follows.

Now the number of positive samples of voxels in ground truth is $S_p$, and the number of negative samples is $S_f$. Denote $\mbox{sigmoid}(\sum_{l=1}^{C_1}\gamma_la_l^i)$ as $\hat{y_i}$, ground truth as $y_i$, and use the BCE loss to calculate the cost.
\begin{equation}
	\mathcal{L}_{\mbox{bce}}(\hat{y_i},y_i)=\sum_{i=1}^{whd}y_i\mbox{log}\hat{y_i}+(1-y_i)\mbox{log}(1-\hat{y_i})\label{bce}
\end{equation}
Then the gradient generated by each voxel is $\eta\frac{\partial\mathcal{L}_{bce}}{\partial \hat{y}_i}$, where $\eta$ is the learning rate. The gradient $\mathcal{G}$ propagated to the next layer is expressed as equation (\ref{f8}).
\begin{equation}
	\mathcal{G} = \eta\sum_{i=1}^{whd}\lambda_i\frac{\partial\mathcal{L}_{\mbox{bce}}}{\partial \hat{y_i}}\label{f8}
\end{equation}
where,
\begin{equation}
	\lambda_i=\begin{cases}
		\frac{S_f}{S_p}\   &\mbox{if} \ Positive\\
		\ 1 \              &\mbox{if} \ Negative\\
		\ 0 \              &\mbox{if} \ Nonlabelled
	\end{cases}
\end{equation}
$\frac{S_f}{S_p}$ is the balance weight coefficient for positive and negative samples. And the $\lambda_i$ is the backward gradient coefficient, so it is equivalent to acting on the loss function to obtain the $\lambda$-BCE loss function.
\begin{equation}
	\mathcal{L}_{\lambda- \mbox{bce}}(\hat{y_i},y_i)= \lambda_i \sum_{i=1}^{whd}y_i\mbox{log}\hat{y_i}+(1-y_i)\mbox{log}(1-\hat{y_i})
\end{equation}

In the same way, we get $\lambda$-smooth $L_1$ loss function.
\begin{equation}
\mathcal{L}_{sL_1}(\hat{\theta}_i,\theta_i)=\lambda_i\sum_{i=1}^{whd}\mbox{smooth}_{L_1}(\hat{\theta}_i-\theta_i)
\end{equation}

Because in the attention map label, the values of the coefficients are continuous, there is no need to calculate balance weight coefficient, we just let the $\lambda_i$ of non-labeled samples be 0.

Although this method also uses 2D slices for training, it is essentially different from 2D seismic fault segmentation. When using this loss function, the input and output are both 3D, and the model fully extracts the three-dimensional features of seismic data, so that it has several times the information of the 2D model when deciding each pixel category. In the experiment compared with 2D seismic fault segmentation, the method in this paper shows significant advantages (Figure \ref{fig10}, \ref{fig11}).

\subsection{Seismic data division and splicing}
Generally, the distribution of faults in seismic data is uniform, so in actual operations, we sample at equal intervals along the inline or crossline directions of the seismic volume and label the sampled 2D slices, and then form the labeled 2D data into the grid. 
Among them, the label of the fault voxel is 1, the non-fault is 0, and the voxels between the two labeled slices are all marked as -1 (unknown voxels). Finally, the seismic volume is divided cuboids  with a size of  $64\times64\times64$.

In the experiment, for real data, we label slices on the original seismic data volume, and then divide  them into cuboids. For synthetic data, the slices in cuboid are directly extracted as labels. We will explain this process in detail in the experiment.

\begin{figure}[htb]
	\includegraphics[scale=0.20]{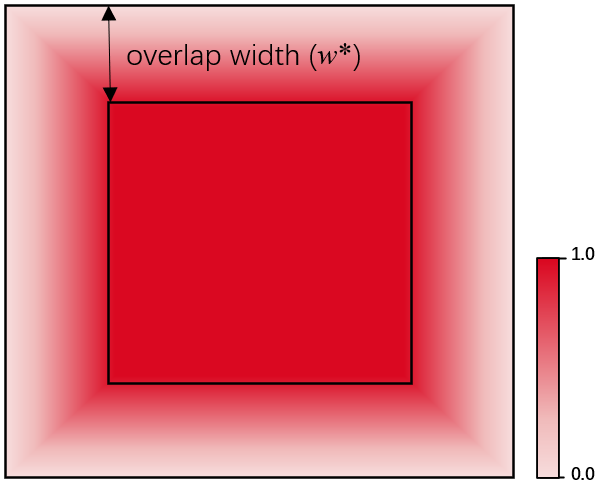}
	\centering\caption{2D visualization of cuboid weights.}
	\label{fig12}
\end{figure}

Making inferences with cuboid with a size of  $64\times64\times64$, and then splicing the cuboids into a seismic data volume of the original size. When splicing, set the overlap width of cuboid to $w^*$, and the weight of the overlap follows the gaussian attenuation from  $\sigma^*=\frac{1}{3}w^*$. As shown in Figure \ref{fig12}, for the convenience of visualization, we show a schematic diagram of 2D weights.

The output of the model is activated by sigmoid function, and the probability is between 0-1. The output of the model is multiplied by the cuboid weight and spliced together, where the probability of the overlap part $\hat{y}^*$ is expressed as.
\begin{equation}
\hat{y}^*=\frac{\sum_{i=1}^{\mathcal{M}}\varphi_i\bar{\hat{y}}_i^*}{\sum_{i=1}^{\mathcal{M}}\varphi_i}
\end{equation}
Where $\bar{\hat{y}}_i^*$ is the probability before splicing, $\varphi_i$ is the weight of cuboid and $\mathcal{M}$ is the number of superimposed cuboids in a position, $\mathcal{M}\in\{1,2,4,8\}$. This splicing method can make the splicing place of each cuboid more smooth.
\section{Experiment}
\subsection{Illustration of the experiment}

\begin{figure}[htb]
	\includegraphics[scale=0.37]{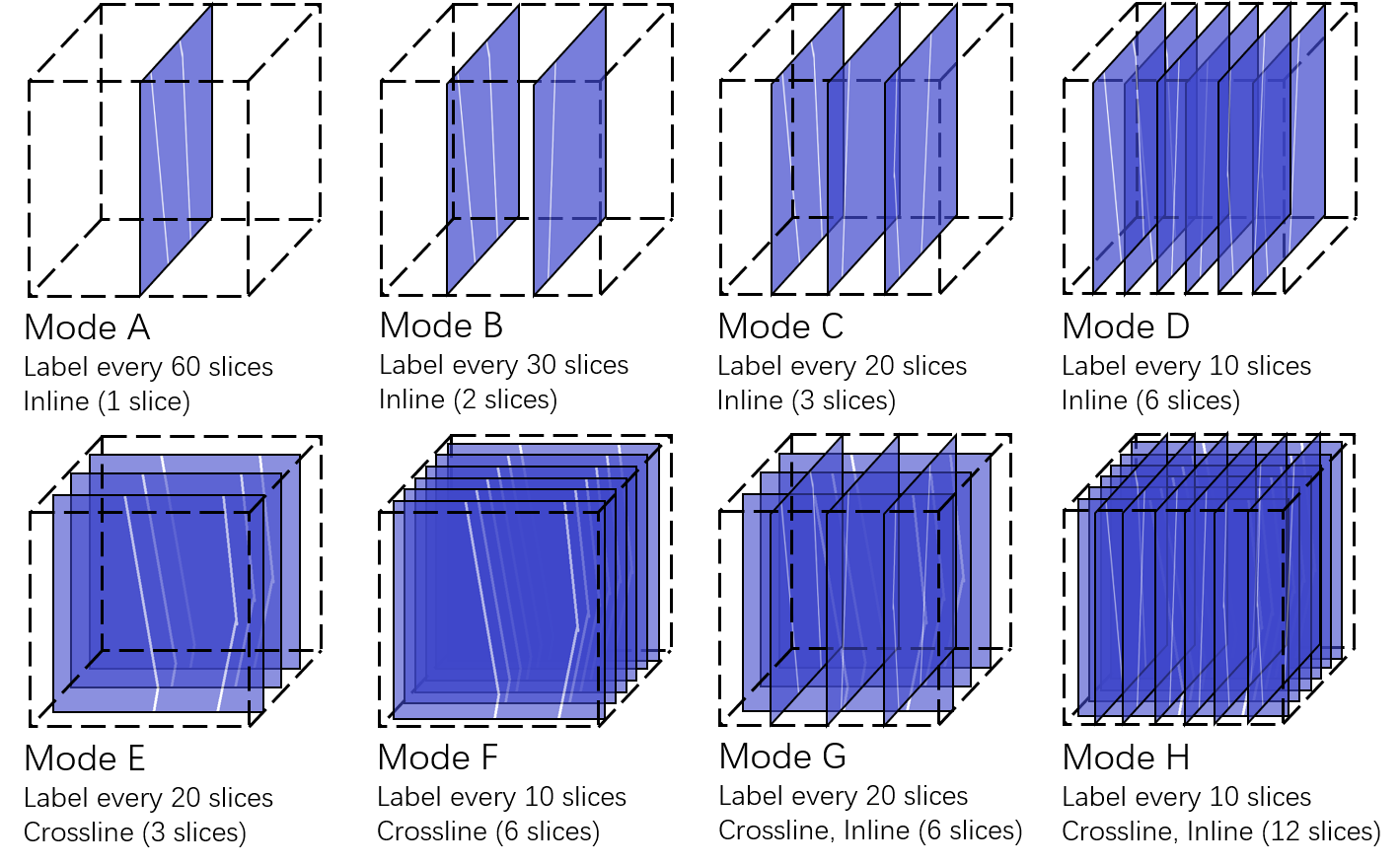}
	\centering\caption{Labeling mode of data.}
	\label{fig4}
\end{figure}

According to the loss function of $\lambda$-BCE and $\lambda$-smooth $L_1$, each sample can participate in training as long as one 2D slice is labeled.
In order to verify the most efficient labeling method, so as to save more labeling costs and help geological professionals to improve efficiency as much as possible, we have verified eight labeling modes, as shown in Figure \ref{fig4}.

\begin{figure*}[ht]
	\includegraphics[scale=0.35]{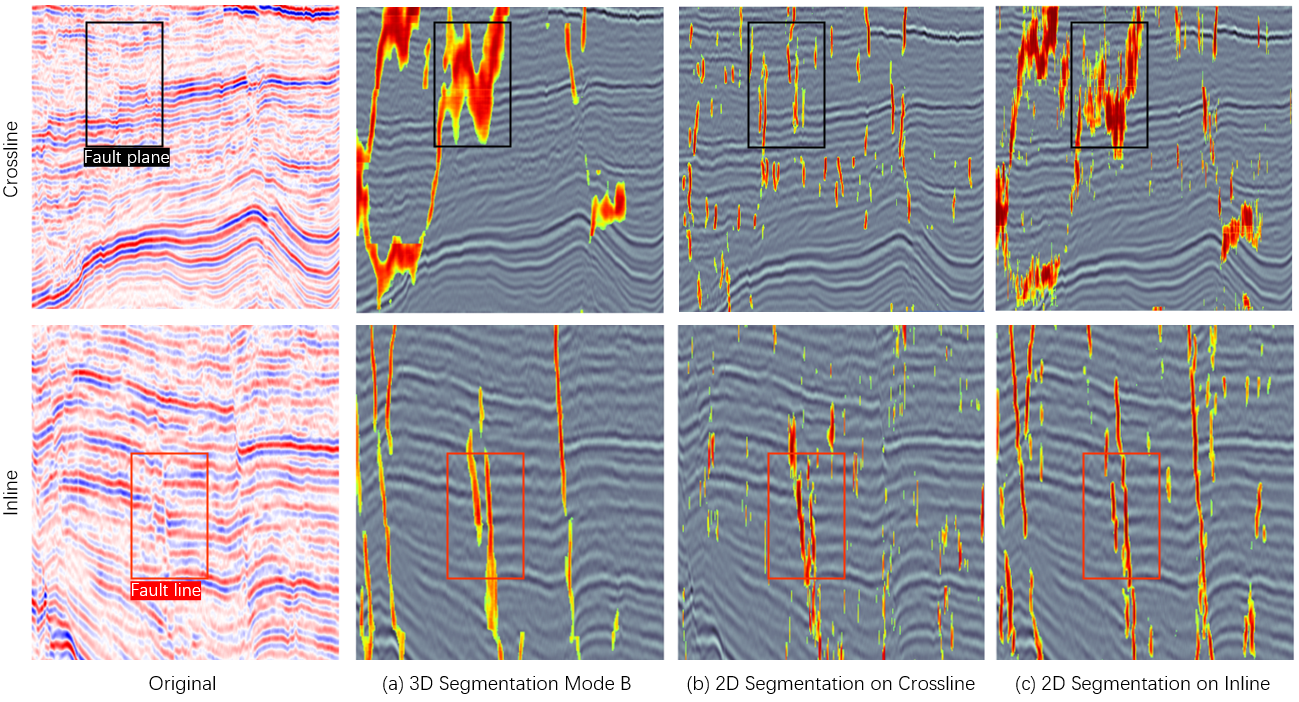}
	\centering\caption{The original data uses pseudo colors. (a) use 3D segmentation. (b) and (c) use 2D segmentation after slicing the data from crossline and inline respectively and combine the segmentation results into the volume. The fault line is perpendicular to the slice, so it is easy to observe and label it, while the fault plane is parallel to the slice, which is very difficult to observe manually.}
	\label{fig10}
\end{figure*}

\begin{figure*}[ht]
	\includegraphics[scale=0.33]{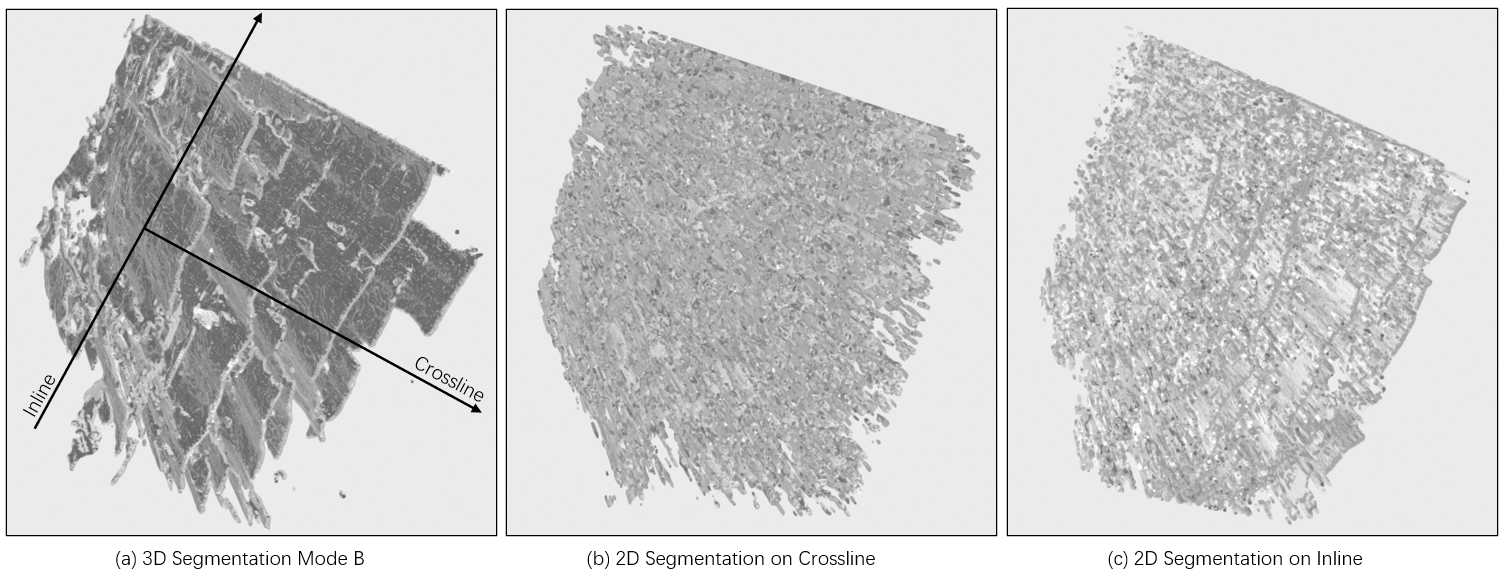}
	\centering\caption{The point cloud reflects the advantages of 3D segmentation. The 3D segmented faults are smooth and clear. The 2D segmentation shows a lot of noise and discontinuous faults.}
	\label{fig11}
\end{figure*}
\begin{figure*}[]
	\centering\includegraphics[scale=0.63]{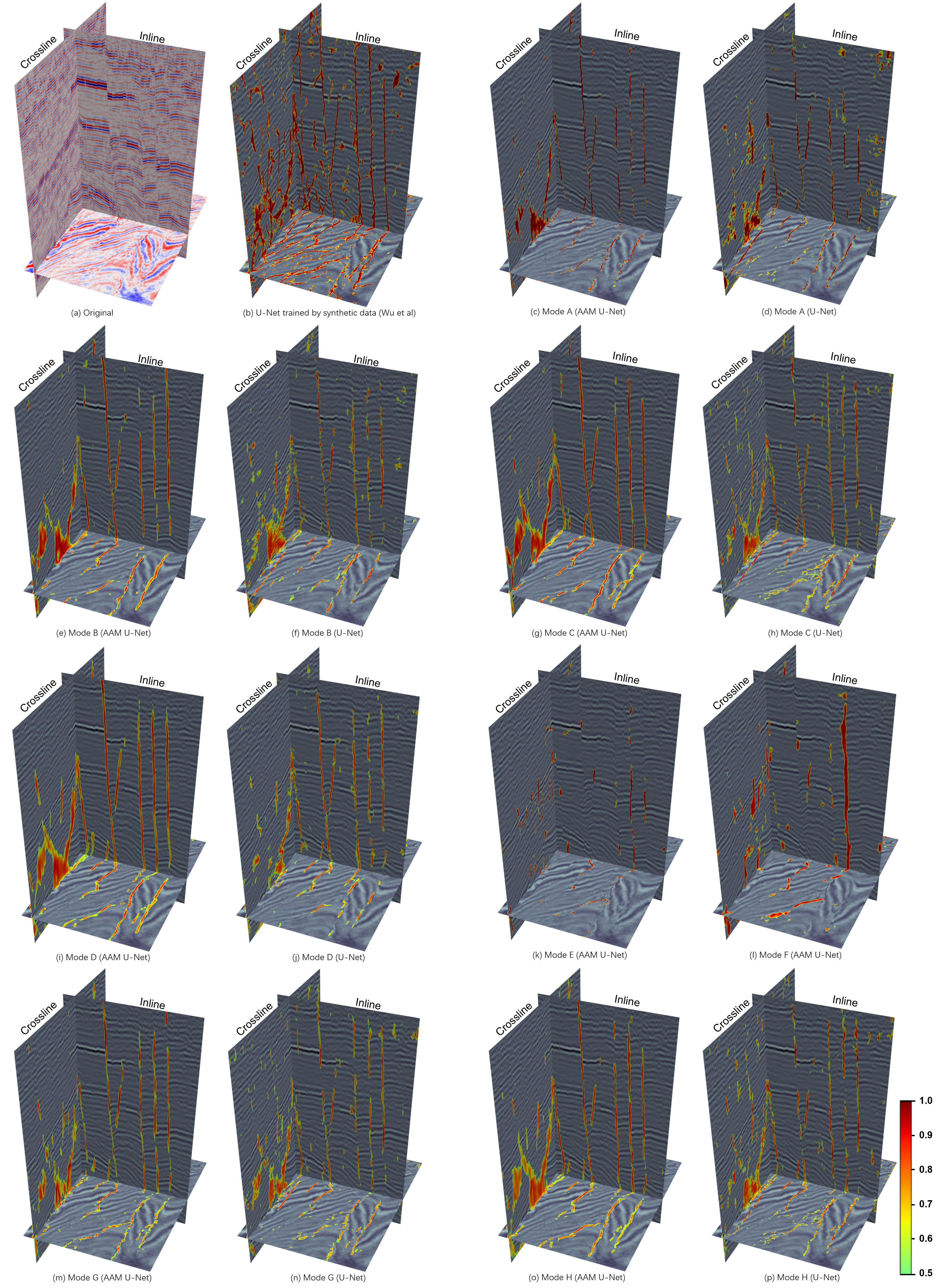}
	
	\centering\caption{Experimental results under training on real data. To facilitate visualization, the original data uses pseudo colors.}
	\label{fig5}
\end{figure*}
\begin{figure*}[htb]
	\centering\includegraphics[scale=0.36]{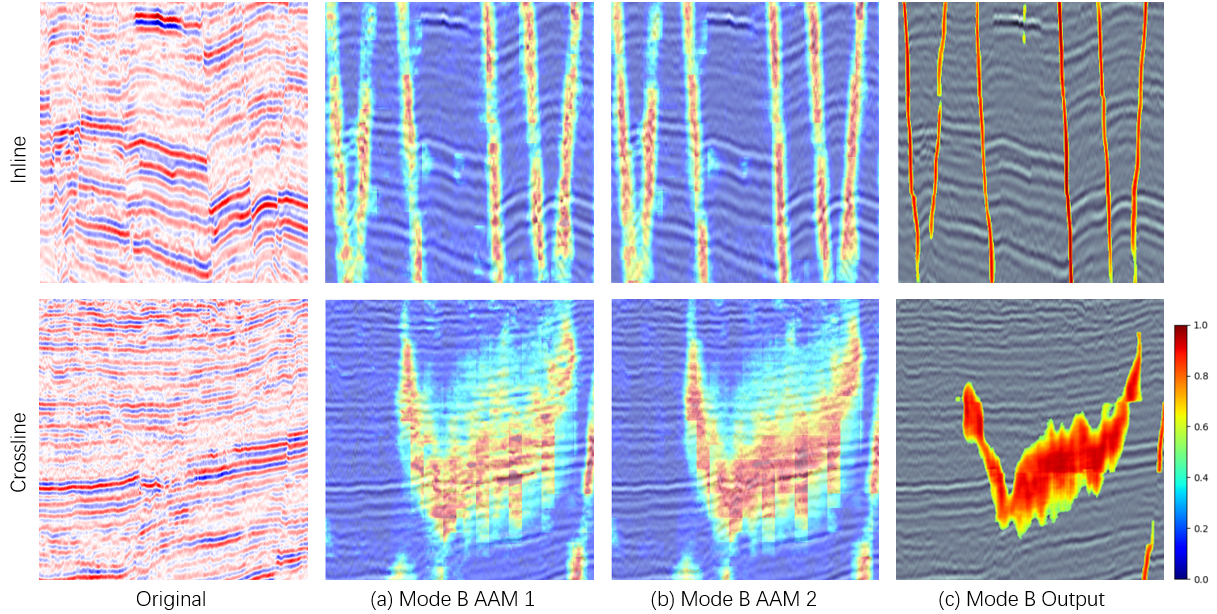}
	\centering\caption{The original data uses pseudo colors. (a) and (b) are the attention map extracted by the two AAM modules in Mode B, respectively. The figure shows that AAM retains the features of the fault and the area around the fault, while suppressing the features of the background area. In crossline, even if we can't observe the fault plane, AAM can still respond to the area in which it is located.}
	\label{fig6}
\end{figure*}
\begin{figure*}[htb]
	\includegraphics[scale=0.38]{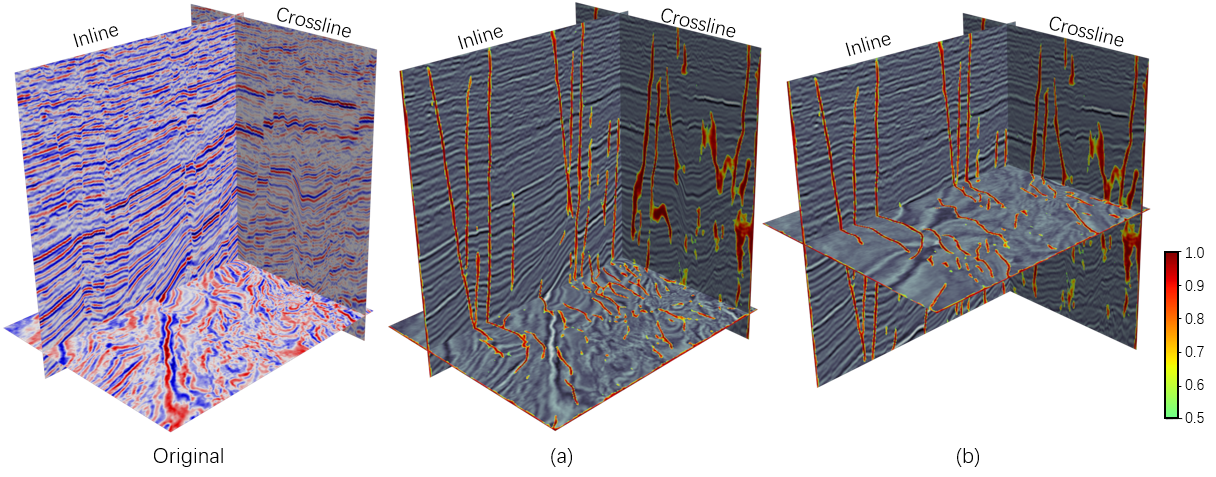}
	\centering\caption{Original seismic data (pseudo color display) and segmentation results of work area $\mathcal{A}$. (a) Bottom timeline slice segmentation results. (b) Middle timeline slice segmentation results.}
	\label{fig16}
\end{figure*}
\begin{figure*}[htb]
	\includegraphics[scale=0.35]{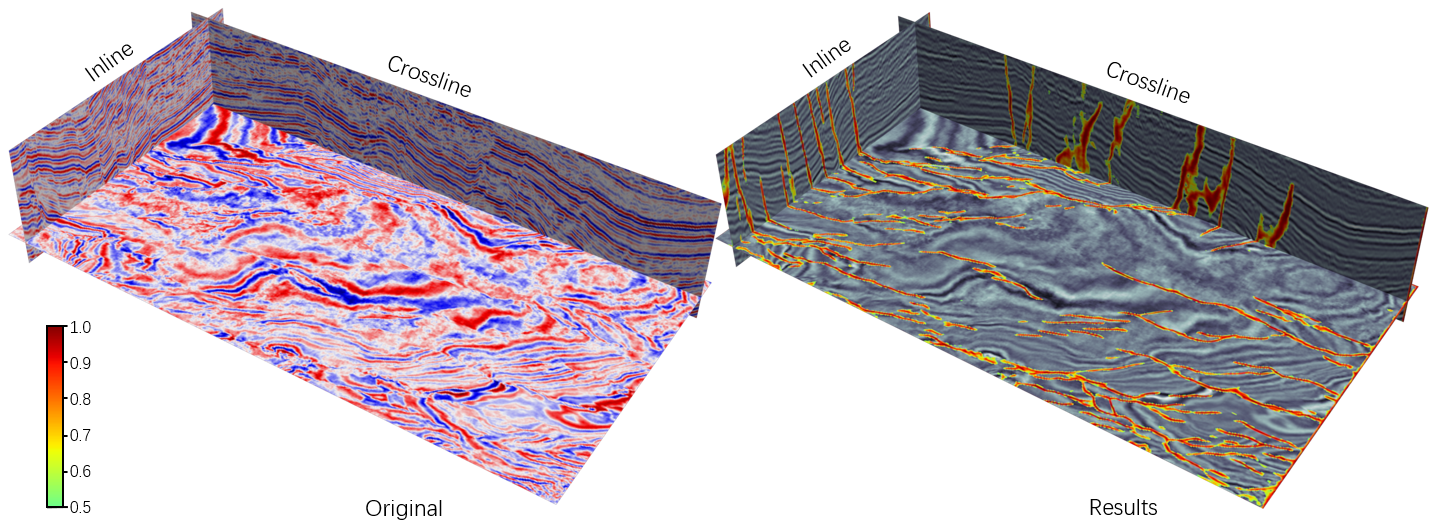}
	\centering\caption{Original seismic data (pseudo color display) and segmentation results of work area $\mathcal{C}$.}
	\label{fig15}
\end{figure*}
\begin{figure*}[htb]
	\includegraphics[scale=0.35]{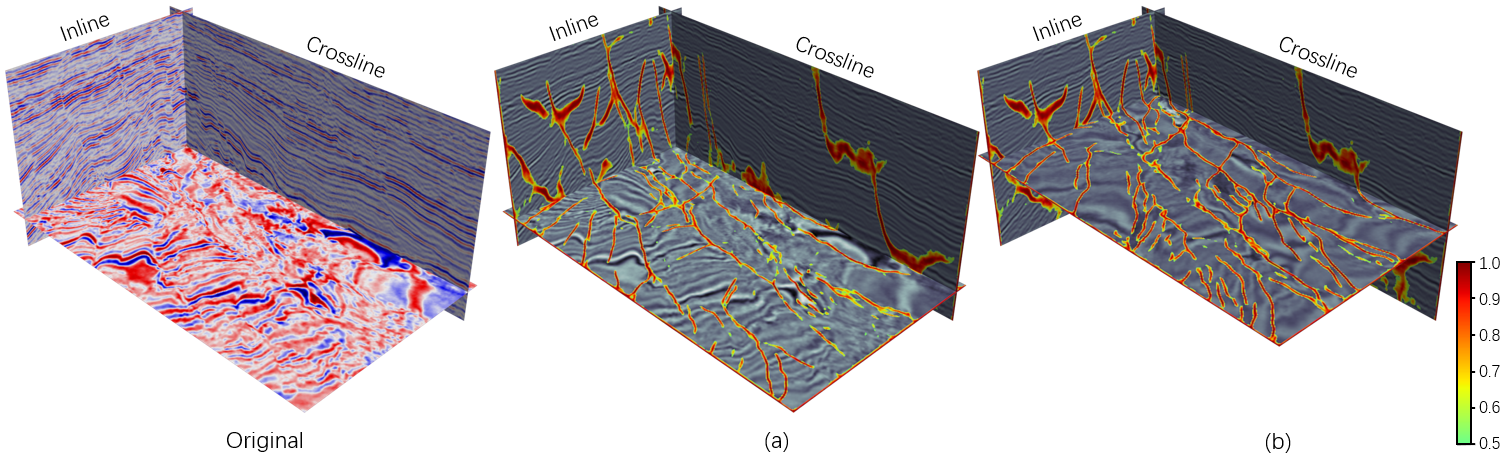}
	\centering\caption{Original seismic data (pseudo color display) and segmentation results of work area $\mathcal{D}$. (a) Bottom timeline slice segmentation results. (b) Middle timeline slice segmentation results.}
	\label{fig18}
\end{figure*}
\begin{figure}[htb]
	\includegraphics[scale=0.35]{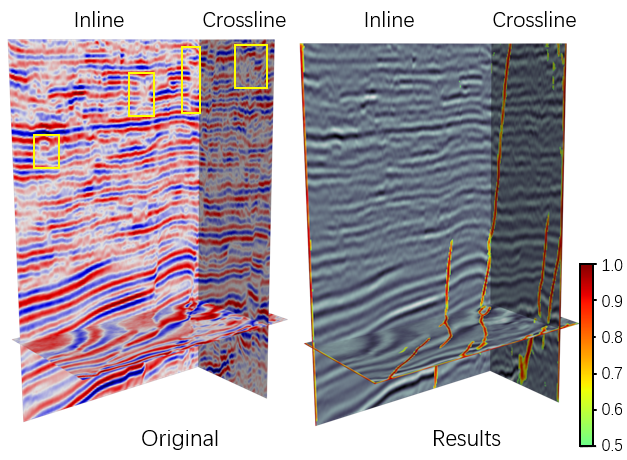}
	\centering\caption{Original seismic data (pseudo color display) and segmentation results of work area $\mathcal{E}$.}
	\label{fig17}
\end{figure}
\begin{figure}[htb]
	\includegraphics[scale=0.32]{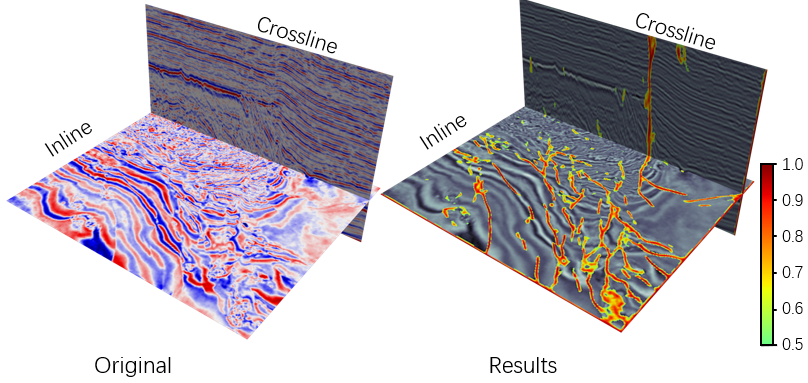}
	\centering\caption{Original seismic data (pseudo color display) and segmentation results of work area $\mathcal{F}$.}
	\label{fig19}
\end{figure}
We conduct experiments on real data and synthetic data to verify our method. Our real data comes from the Shengli Oilfield Branch of Sinopec, and thanks to Wu et al. for published the synthetic seismic fault dataset \cite{22}. There are significant differences in the numerical distribution of different seismic data, so in the experiment we have standardized and normalized all the data.

In real data. First, we compare and analyze the 2D and 3D seismic fault segmentation. Second the influence of different labeling modes on the performance of the model is discussed. Then the effectiveness of AAM in real seismic data is verified. Finally, we have tested and analyzed on several work areas using Mode B to verify that our approach has good generalization capability.

In synthetic data. First, the effectiveness of AAM and the performance of the model in different labeling modes are verified through numerical validation tests. We found that using very few slices training (Mode B) can achieve analogous performance to the 3D volume label, and show the slices and point cloud visualization of the two on the publicly test data. Finally, cross-validation is performed in Mode B.

\subsection{Experiments on real data}

We have data from six work areas ($\mathcal{A}, \mathcal{B}, \mathcal{C}, \mathcal{D}, \mathcal{E}, \mathcal{F}$), where $\mathcal{A}$ is used as the train set and the others as the test set. Annotate the train set as shown in \ref{fig4} and divide it into 5000$\times$64$\times$64$\times$64 samples. The operation is as follows.

The size of our  training data is $1000\times400\times450$, if we use Mode B to label the data.
First, on the inline, label a slice every 30 slices with a size of $1000\times450$. The positive samples are labeled as 1, the negative samples are labeled as 0, and the voxels between the two labeled slices are all marked as -1. Because it is difficult for the eyes to observe fault planes, when labeling, we only label fault lines (see Figure \ref{fig10} for explanations of fault lines and fault planes), and the labeling tool is Labelme \cite{labelme2016}, a seismic volume with size of $1000\times 400\times 450$ only needs to label 12 slices of $1000\times450$ size. Second, slide sampling in three directions, the sliding stride is 25, and the size of each sample is $64\times64\times64$. When sampling, samples with fault voxels less than 64 are deleted.
Finally, the redundant samples are randomly deleted, so that the number of training sets is 5000. Moreover, we label the work area of the test set pursuant to Mode B, and randomly select 500 samples of it as the validation set. 


The experiment used two NVIDIA Tesla P100 16GB (16GB$\times$2 memory), the training epoch is 35 and the batch size is 32. Use Adam optimizer for training \cite{27}, the learning rate is 0.001. We record the $\lambda$-BCE loss on the validation set every 200 steps, and save the current parameters, the  metrics presented in the final experiment are the best among them.
All the experimental codes in this paper are implemented by PyTorch 1.7.0.

Considering that the weights of the 3D convolution kernel are not symmetrical, convolution operations on seismic data from different directions will get different results. If we only label the slice in one direction, it will cause the model to have a serious overfitting problem in this direction. Therefore, we randomly rotate the data during training, which also enables the convolution kernel to fully learn the spatial characteristics of the data.


\subsubsection{Comparison of 3D and 2D seismic fault segmentation}
At present, many works use 2D CNN for semantic segmentation on seismic data slices. There are two main disadvantages. First, 2D segmentation will lose the three-dimensional spatial information of the seismic data, making the segmentation results of adjacent slices discontinuous (Figure \ref{fig10}, \ref{fig11}). Second, the adjacent slices of seismic data are very similar, so they will be sampled and then labeled, and the data between the two sampled slices will be discarded, resulting in a waste of data.

The advantage of 2D segmentation is that there is less computation for seismic data of the same size. FLOPs (floating point operations) \cite{molchanov2016pruning} is used to measure how many floating point operations a model has performed, which usually reflects the inference speed of a model. The FLOPs of the model in this paper for processing cuboid with size of $64\times64\times64$ is 17.35G. After changing all operations of the model to 2D (convolution layer, pooling layer, et al.), the FLOPs of the data of the same size are 0.1716G$\times$64=10.98G (The FLOPs of 64$\times$64 slice processed by 2D is 0.1716G). We train 2D and 3D models with the same number of slices (Mode B), The test results on work area $\mathcal{B}$ are shown in Figure \ref{fig10} and Figure \ref{fig11}, where Figure \ref{fig11} shows the point cloud data converted from 2D and 3D segmentation results.

Figure \ref{fig10} (a) is the result of using our 3D segmentation method. Even if we only label the fault line (just like red box in Figure \ref{fig10}) in the inline when labeling the data, the 3D segmentation results still show a clear and smooth fault plane (just like black box in Figure \ref{fig10}) on the crossline.

But regardless of whether it is segmented in the inline or crossline direction, the 2D model shows extremely poor performance. Especially when segmenting the corssline (Figure \ref{fig10} (b)), because of the lack of 3D spatial information, the model cannot identify the fault plane. In Figure \ref{fig10} (c), although the segmentation result shows fault planes. But its performance is far from 3D model. Even though the calculation of the 3D model is increased by 58\%, the performance improvement is very significant.

In addition, when we observe the  slice of the corssline, it is difficult to observe the existence of fault planes, which leads to many false negative labels when labeling the crossline, it will seriously affect the training process. 

As shown in Figure \ref{fig11}, the inline is often perpendicular to the fault, so the faults in the slice is displayed in the form of a line. Crossline slices are often parallel to faults, and faults are often displayed in the form of planes. It is difficult to observe the fault planes in the slices, and the 2D segmentation model cannot find it in a single slice. Even for human eyes, it is difficult to find fault planes on the crossline parallel to the fault without the help of three-dimensional information. Therefore, when labeling data, label in the inline direction as much as possible.

In our dataset, the labels on the crossline will reduce the performance of the model. We verified this conclusion in the second part of the experiment (Figure \ref{fig5}). But if the label of the crossline slice is accurate, such as in synthetic data, it will not affect performance.

\subsubsection{The effect of train set labeling mode on model performance}

Figure \ref{fig5} displays our method and U-Net, which are trained by work area  $\mathcal{A}$ and tested on work area $\mathcal{B}$. During training, an average of 3.06s elapsed per Adam optimization for our method and 2.65s for U-Net.

The five modes of B, C, G, D, and H show better performance. In these five modes, the increase in the number of slices does not significantly improve the performance of segmentation. We think this is because the adjacent slices of seismic data are very similar, and high-frequency labeling is difficult to increase the diversity of labels.

E and F show a large area of false negative segmentation results, mainly because the training sets of these two modes only label the crossline (parallel to the fault).
As mentioned in previous part, there are many fault planes in crossline slices, which are difficult for humans to observe, which leads to a large number of false negative annotations, which misleads the training process and leads to poor model performance.

However, there may be differences between the data obtained by different work areas or equipment, especially some work areas have crisscross faults. If this type of data is used as a train set, the slice position can be observed from the timeline to avoid fault planes on the slice. But mode G and H also show that when there are enough accurate labels, the model allows a certain number of false negative labels during training.

The method of Wu et al. (U-Net training by synthetic data) has achieved  advanced performance on some data sets \cite{22}, but it has not effectively migrated  to our real data. This demonstrates that it is  difficult to accurately detect faults in real data using only synthetic data  training models.

We tested the impact of different labeling modes on model performance on real data. Experiments and analysis show that in real data, redundant annotations cannot significantly increase the performance of the model. When there are fault planes in the slice (in most cases, it is a crossline slice), the labeled samples may mislead the direction of the propagation process, so try to choose the inline direction to label. A model trained with only synthetic data is difficult to adapt to all types of seismic data.

\subsubsection{Effectiveness of Active Attention Module}

The AAM has a function to suppress noise in the background area. It uses the generated heatmap to suppress the noise introduced during feature fusion. In addition, AAM is also an intermediate supervision mechanism in order to provide more effective gradients.

In Figure \ref{fig5}, the model with the AAM has less noise and better segmentation results. In the  U-Net segmentation results, there will be some false positive noises in non-fault areas, and the addition of AAM can effectively suppress these noises. The main reason is that the attention map generated by AAM retains the area around the fault and suppresses the background area. In Modes B, C, and D, our method shows a higher recall rate. The fault lines detected by the model embedded in AAM are smooth and continuous, while the fault lines detected by the original U-Net have some broken lines. This  indicates that AAM can not only suppress noise, but its intermediate supervision mechanism makes the model more sensitive to the foreground.

Figure \ref{fig6} visualizes the attention map generated by AAM. attention map reflects the spatial distribution of the faults in the slices, especially in the crossline, even if the faults that cannot be manually observed, it can show them significantly.

Experiments on real data show AAM's excellent noise suppression ability, and also show that it can strengthen the model's sensitivity to foreground voxels, thereby increasing recall metrics.

\subsubsection{More tests on seismic data}

To ensure that the model obtained by our method can be more robustly generalized to other seismic data, We tested the performance of the model on five additional work areas ($\mathcal{A}, \mathcal{C}, \mathcal{D}, \mathcal{E}, \mathcal{F}$) using Mode B.

The work area $\mathcal{A}$ is part of the training set in the previous subsection. Although $\mathcal{A}$ is involved in training, only 3.3\% of the slices of it are labeled as supervised information (Mode B), and 96.7\% of the data are automatically fitted by the model, so the test results of $\mathcal{A}$ can reflect the model's ability to fit the original training data. Figure \ref{fig16} is the segmentation result of work area $\mathcal{A}$. On the timeline slices, the model nicely complements the part between the labeled slices, demonstrating continuous and smooth faults due to the $\lambda$ loss function extracting effective gradients from the sparse labels and spreading them throughout the model.

Figure \ref{fig15} shows the segmentation results of the model for the work area $\mathcal{C}$. Most of these faults are perpendicular to the inline, and their fault structure is similar to the training data. The model detects most of the faults, indicating that our method performs good generalization in data with similar faults.

The presence of many crisscross faults in the work area  $\mathcal{D}$ was a great challenge for a model trained using slices in only one direction and labeled only with fault lines, and although the structure of faults in this work area was very different from that of the training data, the network still revealed them very well (Figure \ref{fig18}).
We think it is possible that the network only considers a very limited region in the vicinity of a voxel when deciding its category, i.e., the network is insensitive to the morphology of the fault as a whole in a single sample, and it only focuses on local discontinuities in the seismic data. Which is different from natural and medical images, where deciding the category of each pixel often requires covering the receptive field of the whole image.
Although CNN provide large receptive fields, long-range dependencies are not required for deciding each voxel in the seismic fault segmentation task, the finding that may help in the structural design of CNN for fault segmentation. 
This finding can also be demonstrated in the work of wu et al. Wu's synthetic data incorporates only flat faults, but the resulting trained network still works for curved faults. This is because curved faults are approximated as flat over a very small area.

In Figure \ref{fig17}, the work area  $\mathcal{E}$ shows a very simple fault structure, but there are some fault-like noises in the area marked by the yellow box. Our model cleverly avoids them and gives clean fault segmentation results, showing the excellent noise suppression ability of AAM.

Figure \ref{fig19} displays the segmentation result of work area $\mathcal{F}$. From the inline slice, the work area has a major fault, and its left side is full of broken reflectors. 
Although this region has a low signal-to-noise ratio and has significant discontinuities, our method does not show much noise in this region, but instead decides it all as background. 
Therefore the discontinuity of reflectors is only used as one of the bases when the model segmented the fault, it also provides more advanced semantic features to evade false positives. 

In this subsection, we perform further tests on five work areas using the model trained by slices representing 3.3\% of the real data volume. The test results demonstrate the ability of our method to generalize to multiple types of seismic data with only a few slice labels trained.

\subsection{Experiments on synthetic data}
\subsubsection{Ablation experiment}

Real data requires manual labelling of faults, so it is difficult to avoid incorrect labeling. The label of synthetic data is automatically generated based on an algorithm. It is completely accurate, so its performance numerical metrics on the validation set can reflect the performance of the model very accurately. Therefore, we designed an ablation experiment based on synthetic data to verify the effectiveness of AAM. The ablation experiment is a controlled variable method that verifies its effectiveness in the model by adding or deleting a component\cite{26}.

We split the data of size $128\times128\times128$ published by Wu into 8 pieces of data of $64\times64\times64$. In addition, we also downsample (resize) each original data into size of  $64\times64\times64$, a total of $220\times(8+1)=1980$ data, 300 of which are randomly sampled as validation set. We use the six modes shown in Figure \ref{fig4} to process labels (The synthetic data cannot distinguish between crossline and inline, so C and E are regarded as the same mode, and so are D and F), and the 3D volume label (all label). A total of seven label modes for training. The experiment uses the Adam optimizer, the learning rate is 0.001, and the epoch is 100. We record the $\lambda$-BCE loss on the validation set every 200 steps, and save the current parameters, the  metrics presented in the final experiment are the best among them.

In the experiment, we found that the data augmentation method of randomly rotating samples is very effective for training, but it is often ignored when used in 3D tasks. We added this to the ablation experiment. Therefore, our ablation experiment contains two variables: whether to add AAM and whether to rotate the sample cuboid during training. It should be noted that the Non-AAM model is the original U-Net, which is also the model used by Wu et al. \cite{22}.

We use IOU (Intersection Over Union) as the performance evaluation metric. The IOU is expressed by equation \ref{iou}.
\begin{equation}
	IOU=\frac{TP}{FP+TP+FN}\label{iou}
\end{equation}
Among them, TP (True Positive) is classified as a positive sample, in fact it is also a positive sample. FP (False Positive), is classified as a positive sample, but in fact it is a negative sample.
FN (False Negative) is classified as a negative sample, but in fact it is a positive sample. 
The output of the model is normalized to the probability $\hat{y}_i\in[0,1]$ through the sigmoid activation function, and $\hat{y}_i>0.5$ is regarded as a positive sample, $\hat{y}_i\leq0.5$ is regarded as a negative sample, so as to calculate the IOU.

\begin{table*}[ht]
	\caption{Ablation Experiment}
	\label{t1}
	\centering

	\resizebox{\textwidth}{!}{%
		\begin{tabular}{cccccccccc}
			\toprule
			AAM & \begin{tabular}[c]{@{}c@{}}Random\\ Rotation\end{tabular} & \begin{tabular}[c]{@{}c@{}}All Label\\ 64,64 slices\end{tabular} & \begin{tabular}[c]{@{}c@{}}Mode H\\ 6,6 slices\end{tabular} & \begin{tabular}[c]{@{}c@{}}Mode G\\ 3,3 slices\end{tabular} & \begin{tabular}[c]{@{}c@{}}Mode D/F\\ 6 slices\end{tabular} & \begin{tabular}[c]{@{}c@{}}Mode C/E\\ 3 slices\end{tabular} & \begin{tabular}[c]{@{}c@{}}Mode B\\ 2 slices\end{tabular} & \begin{tabular}[c]{@{}c@{}}Mode A\\ 1 Slice\end{tabular} & FLOPs (G) \\ \midrule
			&                                                           & 69.72                                                            & 63.04                                                       & 65.59                                                       & 60.48                                                       & 55.10                                                        & 37.44                                                     & 33.60                                                     & 17.01     \\
			$\surd$   &                                                           & 70.10                                                             & 69.15                                                       & 67.66                                                       & 61.34                                                       & 58.64                                                       & 39.48                                                     & 36.56                                                    & 17.35     \\
			& $\surd$                                                         & 71.43                                                            & 70.22                                                       & 68.19                                                       & 66.88                                                       & 64.67                                                       & 64.17                                                     & 59.59                                                    & 17.01     \\
			$\surd$   & $\surd$                                                         & 72.18                                                            & 71.69                                                       & 70.01                                                       & 70.92                                                       & 69.65                                                       & 68.83                                                     & 64.41                                                    & 17.35     \\ \bottomrule
		\end{tabular}%
	}
\end{table*}

\begin{figure}[htb]
	\centering\includegraphics[scale=0.3]{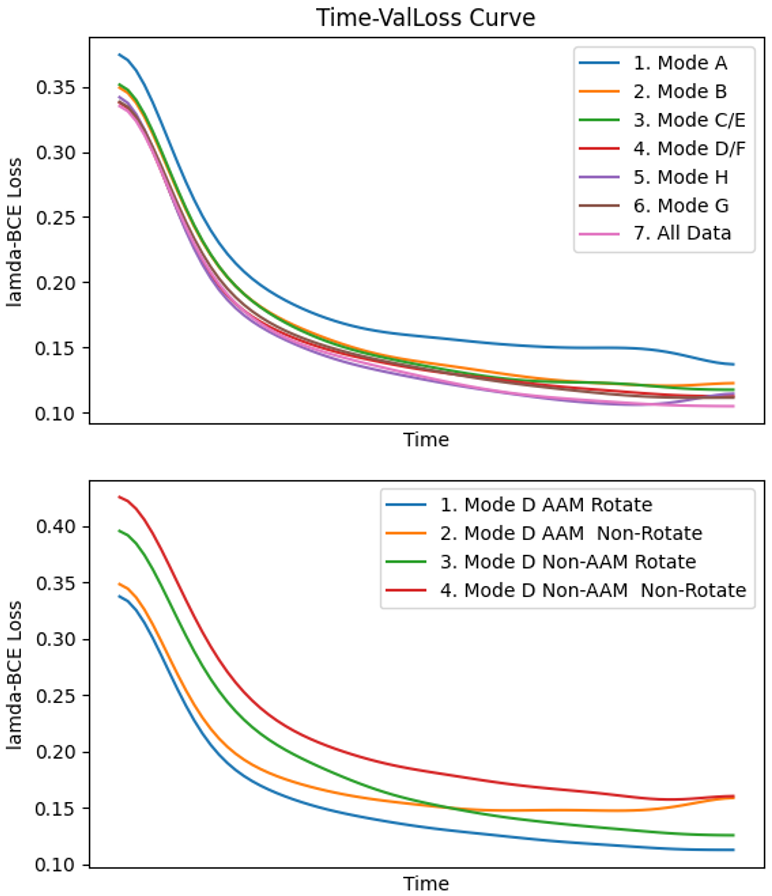}
	\centering\caption{In order to facilitate observation, the curve has been smoothed. (a) shows the convergence curve of the model after using AAM and random rotation of the sample. When the labeled slice is greater than 1, the convergence of the model is similar.
	(b) shows when an ablation experiment is performed in Mode D (6 slices). The figure shows that the performance of the non-rotating model quickly reaches saturation and gradually declines (overfitting). It also shows that using AAM will make the model converge faster and have a higher upper limit.}
	\label{fig8}
\end{figure}
\begin{figure*}[]
	\centering\includegraphics[scale=0.72]{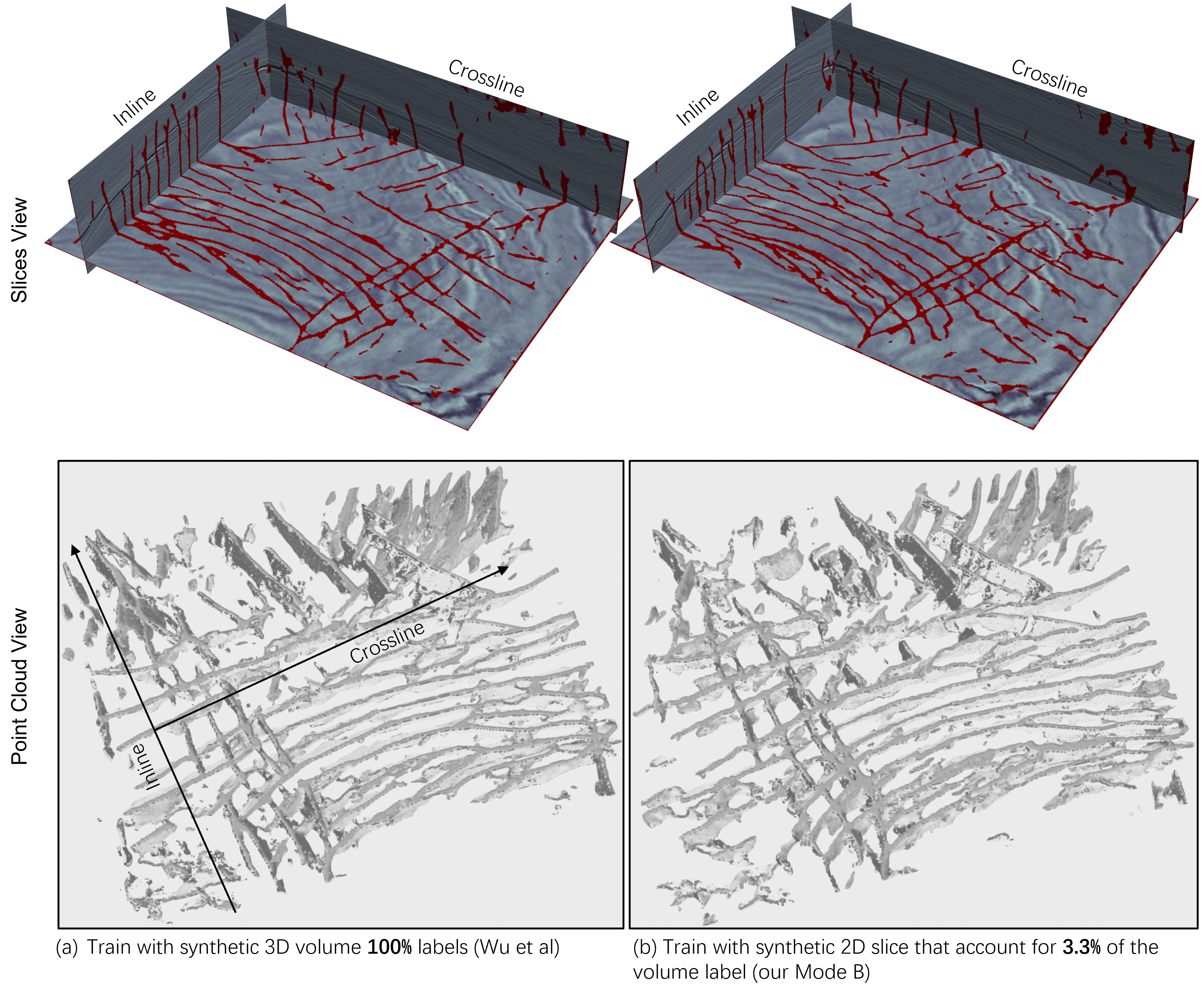}
	\centering\caption{As shown in the ablation experiment, the IOU of Mode B under our method is 68.83, while the IOU of the original U-Net trained with all label is 70.1. Although the number of labels trained by the latter is 30 times that of the former, its performance has not been significantly difference.}
	\label{fig13}
\end{figure*}
\begin{figure*}[]
	\centering\includegraphics[scale=0.72]{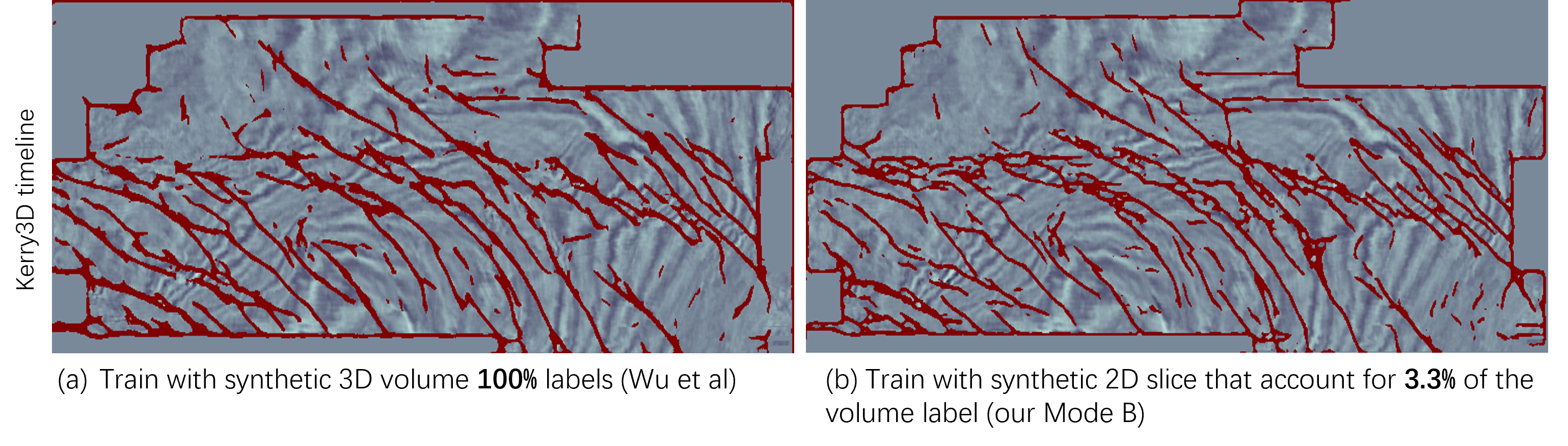}
	\centering\caption{Timeline slice of Kerry3D, the experimental results again show similar performance between the two.}
	\label{fig14}
\end{figure*}
Figure \ref{fig8} shows the loss of the validation set over time. From Table \ref{t1}, it shows that AAM can improve	 numerical metrics. Although there is weak noise in the synthetic data, the use of AAM still improves the performance of the model.

It is worth noting that during training, whether or not to rotate the sample plays a decisive role in the performance of the model. Especially when the sample is only labeled in one direction, if the sample is not rotated, the performance of the model will be greatly reduced.

There is no noticeable difference in the numerical performance metrics between All Label and Model B-H.
This phenomenon is also reflected in Figure \ref{fig5}. High-frequency labeling slices will not significantly improve training performance. It may be unnecessary to consume too much labor cost to label seismic data.

Finally, we tested the performance of the model on public data.
The data in Figure \ref{fig12} comes extracted from the Netherlands off-shore F3 block seismic data. which is graciously provided by the Dutch government through TNO and dGB Earth Sciences \cite{22}. The data in Figure \ref{fig13} is Kerry-3D, which is provided on the SEG Wiki website.

We show the results of segmentation using Mode B (AAM U-Net trained with 3.3\% synthetic data label), and compare it with the method of Wu et al. (original U-Net trained with 100\% synthetic data label).

In Figure \ref{fig13}, there is no significant difference between the two in the slice view. The 3D point cloud view shows that although our method is trained with very few 2D slices, the segmented faults are as smooth and continuous as the method of Wu et al.
In Figure \ref{fig14}, the comparison of the slices in the Kerry3D timeline also shows that there is no significant difference between the two. This may mean that the training of the 3D seismic segmentation model does not demand  complete 3D volume labels, which contain a lot of repetitive and redundant information.

We found that in synthetic data without strong noise, AAM can still significantly improve model performance (Table \ref{t1}).
This is because AAM provides an intermediate supervision mechanism, which allows the model to obtain more effective gradients during training. 

We verified that the seismic data obtained for a certain work area or a certain instrument, we only need to label  once every 30 slices to make the model obtain very good segmentation performance.
Next, we use cross-validation to further confirm our conclusion.
\subsubsection{Cross validation}
This experiment uses the K-fold cross-validation method. We set K=5, and divide the 1980 samples into 5 sub-samples evenly and randomly, a single sub-sample is kept as the validation set, and the other 4 sub-samples are used for training. The cross-validation was repeated 5 times, and each sub-sample was validated once. The advantage of this method is that random sub-samples are repeatedly used for training and verification simultaneously, which means that each sample will be used for verification. Among them, all training data uses Mode B (2 slices labeled, i.e. labeled once every 30 slices).

The cross-validation experiment uses five metrics: Precision, Recall, IOU, Dice and Hausdorff Distance.
Among them, Precision and Recall are common metrics in machine learning, IOU and Dice are a metric that is sensitive to the segmentation area, and Hausdorff Distance is a metric that is sensitive to the segmentation boundary. For more analysis of these five metrics, see reference \cite{28}.

\begin{table}[htb]
	\caption{Cross Validation for Two Slices Labeled}
	\label{t2}
	\centering
		\begin{tabular}{ccccccc}
			\toprule
			& \textbf{\textit{Precision}} & \textbf{\textit{\ Recall\  }} & \textbf{\textit{\ IOU\ }} & \textbf{\textit{\ Dice\ }} & \textbf{\textit{Hausdorff}} \\ \midrule
			\textbf{\textit{set 1}} & 76.58              & 89.30           & 66.86        & 80.14         & 62.84              \\
			\textbf{\textit{set 2}} & 76.07              & 88.47           & 65.97        & 79.50         & 66.43              \\
			\textbf{\textit{set 3}} & 75.88              & 91.40           & 66.51        & 79.88         & 69.29              \\
			\textbf{\textit{set 4}} & 75.74              & 89.17           & 65.74        & 79.33         & 65.80              \\
			\textbf{\textit{set 5}} & 77.06              & 90.25           & 67.78        & 80.80         & 68.16              \\
			\textbf{\textit{Mean}}  & 76.27              & 89.72           & 66.57        & 79.93         & 64.30              \\ \bottomrule
		\end{tabular}%
\end{table}
The cross-validation result data in Table \ref{t2} shows that each sample only uses two slices labeled for training to obtain a very stable and usable model.
The experiment showed a high recall rate and IOU of the model, indicating that the model detected almost all faults.
The reason why precision is slightly lower than the recall is that the width of the fault label in the label is too narrow, and the detected fault is wider, which leads to a slightly larger FP.
The stability of the hausdorff distance above 60 indicates that the model is very advanced in processing the boundary and noise.

\section{Conclusion}
So as to suppress the strong noise in real seismic data, an attention module AAM is proposed and embedded in the original U-Net model. It generates an attention map through active supervised training to suppress the background area of the seismic image. Experiments on real seismic data show that the noise in its segmentation results is significantly lower than the original U-Net. In addition, as an intermediate supervision mechanism, it can also improve the sensitivity of the model to the foreground and can also improve the numerical indicators even on synthetic data with weak noise.

In order to solve the difficulty of labeling in 3D seismic data fault segmentation tasks, $\lambda$-BCE and $\lambda$-smooth $L_1$  loss are proposed, so that we can use very few 2D slice labels to train 3D seismic fault segmentation models. Experiments have shown that our model has significant advantages compared with the 2D segmentation model when using the same 2D slice label. We achieved good generalization over six work areas using only 3.3\% of the slice labels of one work area.
Moreover, with our approach, 2D slice labels using only 3.3\% of the synthetic data achieve similar performance to that using 3D labels.
This not only shows the effectiveness of our method, but also indicates that when training 3D seismic segmentation models, it is not necessary to intensively label the slices of the same work area (no need for dense labels), and there is a lot of redundant and repeated information in the seismic data. This may help geological researchers or workers to use deep learning for seismic fault segmentation or other seismic data segmentation tasks without the need for dense or volume labeling.

\section*{Acknowledgments}
The authors are very indebted to the anonymous referees for their critical comments and suggestions for the improvement of this paper. This work was also supported by grants from the National Natural Science Foundation of China (Major Program, No.51991365).


%

\ifCLASSOPTIONcaptionsoff
  \newpage
\fi



%

\bibliographystyle{IEEEtran}
\bibliography{mybibfile}

\end{document}